%% file: Template.tex
\documentclass{article}
\usepackage{pifont}
\usepackage{spconf,amsmath,graphicx}
\usepackage{amsmath, amssymb}
\usepackage{fancyhdr}
\usepackage{url}
\usepackage{enumitem}
\usepackage{lipsum} % ダミーテキストのためのパッケージ
\usepackage{bm}

\title{Map-Mono-Ego: Map-Grounded Global Human Pose Estimation from Monocular Egocentric Video}
\name{
\begin{tabular}{c}
Hiroyuki Deguchi$^{1,2}$, Ryosuke Hori$^{1,2}$, Kotaro Amaya$^1$ \\
Tsubasa Maruyama$^2$, Mitsunori Tada$^2$, Hideo Saito$^1$
\end{tabular}
}
\address{$^1$Keio University~~$^2$National Institute of Advanced Industrial Science and Technology}
\usepackage{hyperref}
\usepackage{url}
\usepackage{caption}
\usepackage{amsmath}
\usepackage{xcolor}
\begin{document}
%\ninept
%
% \maketitle
%
\newcommand{\Ours}{Map-Mono-Ego}
\newcommand{\Ourdataset}{AIST-Living dataset}
\newcommand{\etal}{\textit{et al.}}
\newcommand{\red}[1]{\textcolor{red}{#1}}
\newcommand{\darkgraybox}[1]{\colorbox[rgb]{0.8,0.8,0.8}{#1}}
\newcommand{\dingc}[1]{\raisebox{-0.1ex}{\scalebox{1.1}{{\ding{#1}}}}}
\twocolumn[{%
\renewcommand\twocolumn[1][]{#1}%
\maketitle
}]

\begin{abstract}
\input{sections_template/0_abst}
\end{abstract}
%もう一個キーワード追加できる

\section{Introduction}
\label{sec:intro}
\input{sections_template/1_intro_v2}

% \input{figs/overview}

% \twocolumn[\input{figs/teaser}]

\section{RELATED WORKS}
\label{sec:related}
\input{sections_template/2_Related}

% \section{BACKGROUND}
% \label{sec:background}
% \input{sections/3_background}

\section{METHOD}
\label{sec:method}
\input{sections_template/3_method_v2}
% ---------result
\begin{figure*}[t]
  \centering
  \includegraphics[width=0.9\linewidth]{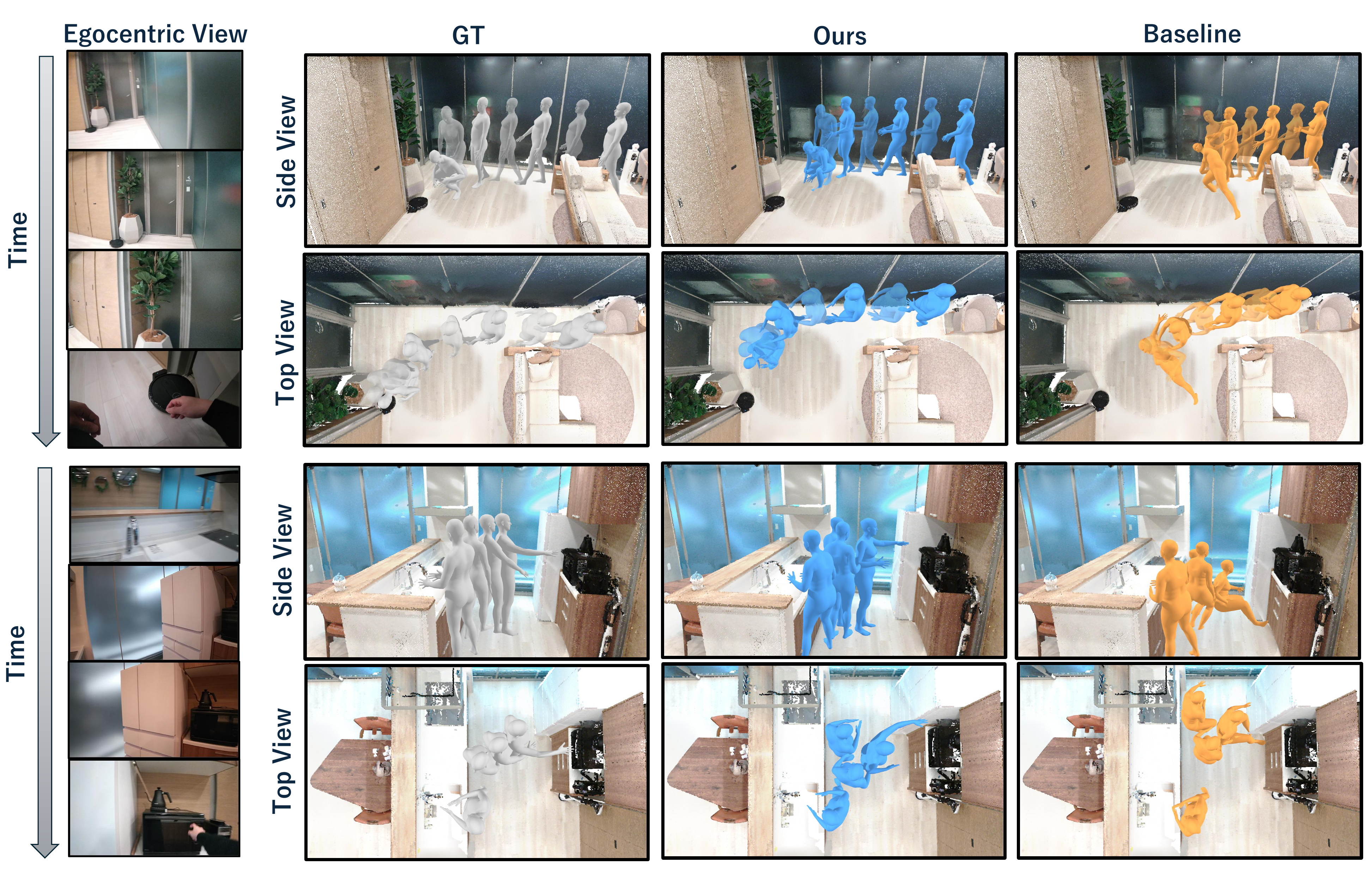}
  \vspace{-8pt}
  \caption{Qualitative comparison of global human pose estimation. The top one shows a sequence where the subject walks to a robotic vacuum cleaner and crouches down. The bottom one shows a sequence where the subject walks to a microwave in the kitchen and reaches for it. Our method estimates more precise and natural motion than the baseline in both sequences.}
  \label{fig:qualitative}
\end{figure*}
% ----------result

\section{EXPERIMENTS}
\label{sec:experiments}

\input{sections_template/4_experiments}

\section{CONCLUSION}
\label{sec:conclusion}

\input{sections_template/5_conclusion}

\vfill\pagebreak
\clearpage
% \section{REFERENCES}
% \label{sec:refs}

\bibliographystyle{IEEEbib}
\bibliography{refs}

\end{document}

%\ninept
%
\maketitle

\hypersetup{linkcolor=black}
\tableofcontents
\hypersetup{linkcolor=red}

\section{Overview of the Supplementary Material}

The supplementary material includes details on implementation and the original dataset. In addition, we show the limitations and additional visual analysis on ablation study of our method.
% Furthermore, we provide a video demo to obtain more qualitative results.

\section{Implementation Details}
\begin{figure}[t]
  \centering
  \includegraphics[width=1.0\linewidth]{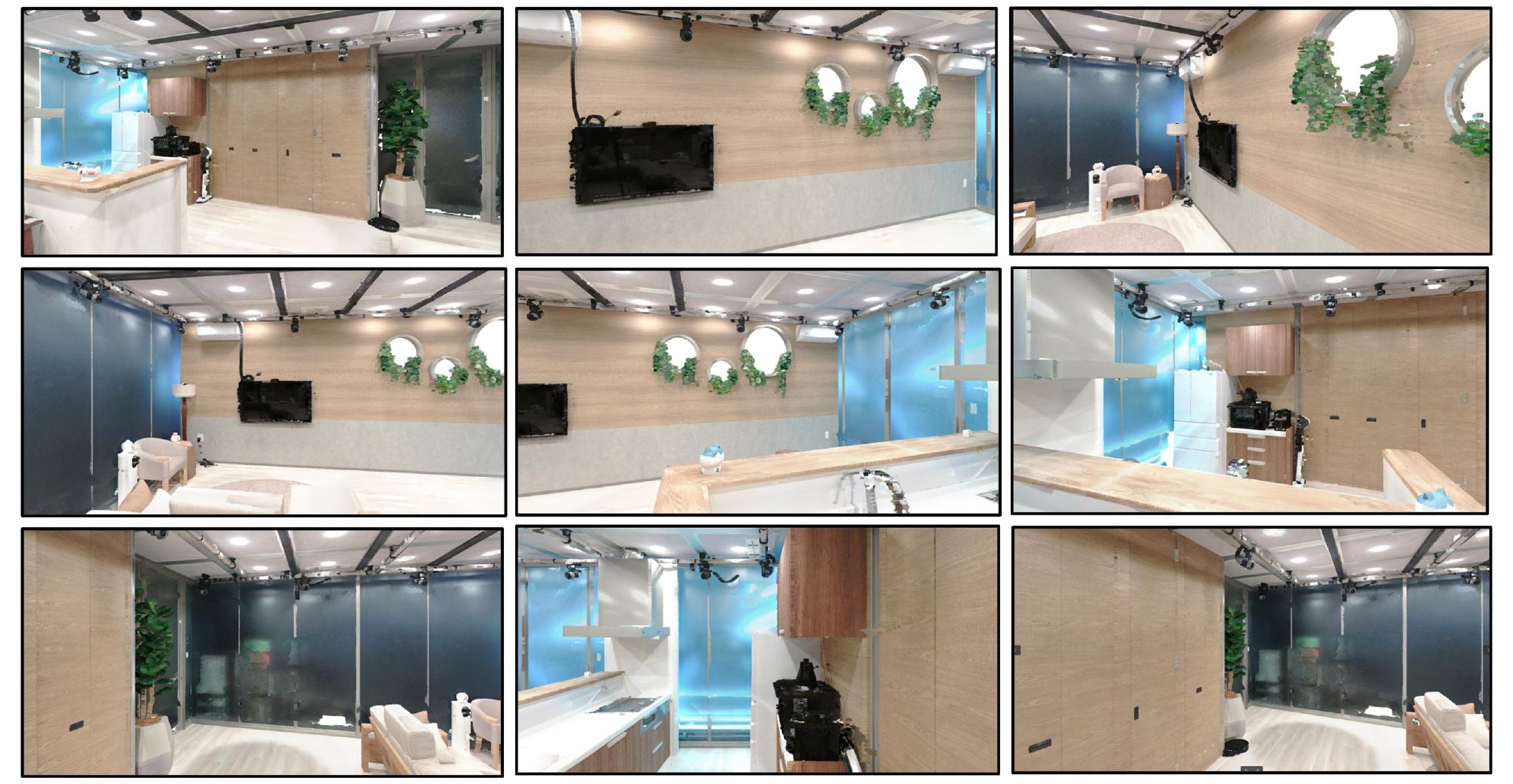}
  \caption{Examples of the synthetic database images.}
  \label{fig:synth}
  \vspace{-15pt}
\end{figure}

% Hlocの特徴量検出器，マッチング，検索手法についても書く
% Inlier率のthresholdとかも書く
% Map上に一定間隔ってどのくらい？
\noindent
\textbf{Localization via synthetic database~ }
To obtain synthetic database, we sampled virtual cameras within the metric point cloud using a grid spacing of 0.15m in the xy-plane and 0.25m along the z-axis (ranging from 0.5m to 1.75m).
While the camera orientation was randomized around the camera's pitch, we discarded positions within a 0.2m distance from the nearest point cloud surface to ensure valid synthetic views and avoid occlusion. For a better understanding of the synthetic database images, Fig.~\ref{fig:synth} shows examples of the synthetic rendered database images. We can obtain such clear synthetic images owing to the high density point cloud caputured by a terrestrial laser scanner.
For HLoc~\cite{hloc}, we employed ALIKED~\cite{ALIKED} for feature extraction, LightGlue~\cite{lightglue} with ALIKED features for matching, and NetVLAD~\cite{netvlad} with top-40 retrievals for global descriptors.

\noindent
\textbf{Trajectory Refinement by Inlier-based Filtering~ }
We identified reliable anchor frames based on an inlier ratio threshold of 50\% and a minimum inlier count of 500.
To facilitate smooth trajectory interpolation via SLAM, we enforced a minimum interval of 20 frames between selected anchors.

\noindent
\textbf{Diffusion-based Human Pose Estimation~ }
We trained the motion diffusion model using a single NVIDIA RTX 3090 GPU.
While we modified the conditioning input to accept neck joint trajectories to align with our experimental setup, all other training hyperparameters, including the number of epochs, optimizer, and learning rate, followed the original UniEgoMotion~\cite{uniegomotion} configuration

% -------------synthetic rendering
% ------------------synthetic rendering

% 手とかが若干難しいとかもここで言及してよさそう
\noindent
\textbf{Baseline~ }
We estimated the baseline trajectory by feeding DROID-SLAM ~\cite{teed2021droid} results into the pretrained GravityNet and HeadNet models.
For HeadNet, we selected the model pretrained on the real-world GIMO dataset~\cite{gimo}, as it yielded the best performance.
Since these models are designed for 30 FPS sequences, we computed the trajectory at 30 FPS and downsampled the output to 10 FPS for evaluation.

\section{Dataset Details}
% 画素数とかも言及する
We captured the original dataset, which pairs environmental point clouds, egocentric video, and ground-truth motion data. We obtained these data by the way as follows.
The static 3D environment was captured using a FARO Focus laser scanner ~\cite{faro} to obtain an accurate and dense point cloud.
Simultaneously, subjects performed common daily living activities while wearing a neck-mounted camera ~\cite{thinklet} to record egocentric video at a resolution of $1920 \times 1080$. To ensure geometric accuracy for visual localization, we performed camera calibration to estimate the intrinsic parameters and compensate for lens distortion.
To obtain ground truth motion in the SMPL-X model ~\cite{smplx}, we processed motion data captured by Theia3D~\cite{theia} through DhaibaWorks~\cite{DW} for synthetic marker generation and Soma~\cite{soma} for model fitting.

% -------------limitation
\begin{figure*}[t]
  \centering
  \setlength{\belowcaptionskip}{-10pt}
  \includegraphics[width=0.8\linewidth]{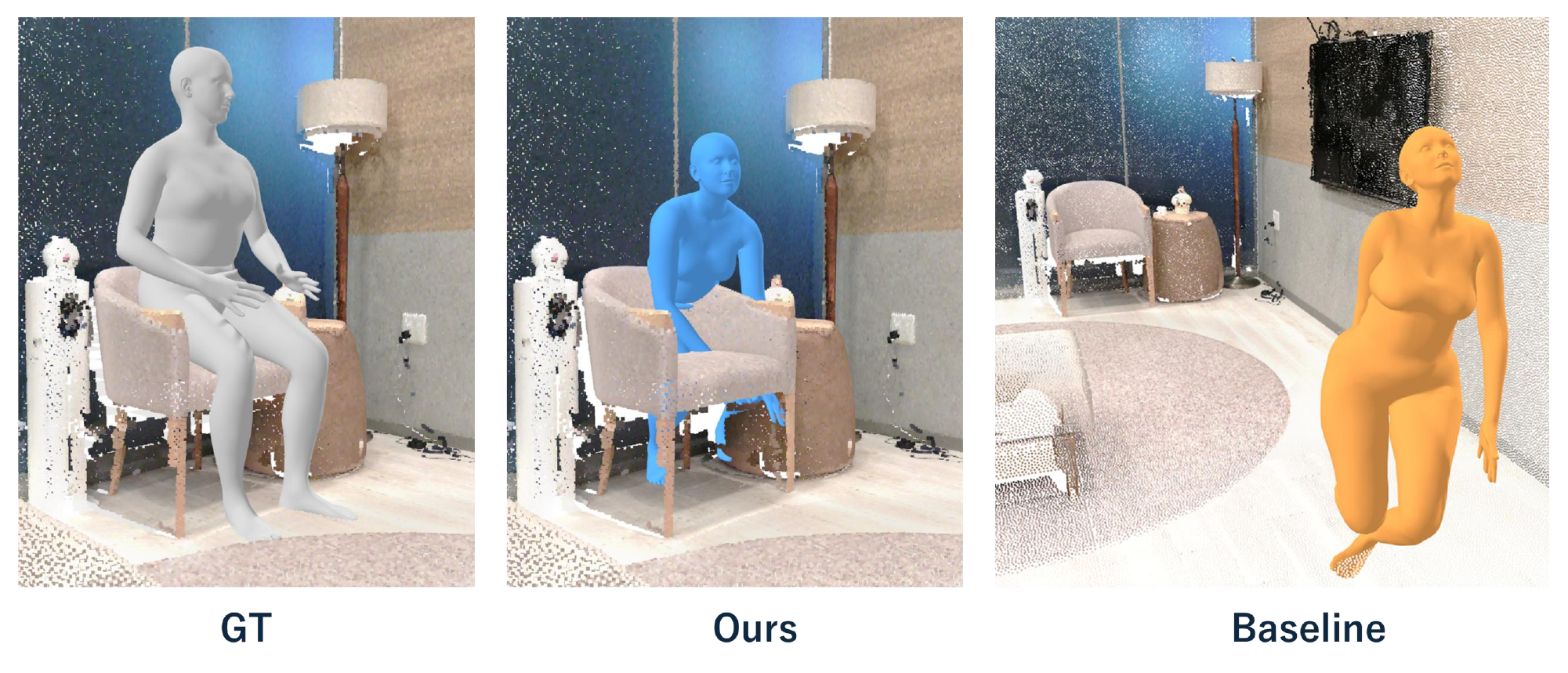}
  \vspace{-10pt}
  \caption{A qualitative comparison between GT, Ours, and Baseline in a sitting frame. While our method estimates reasonable global location and natural sitting posture, the lack of explicit physical constraints leads to unnatural surface penetration artifacts.}
  \label{fig:limit}
\end{figure*}
% --------------limitation

\section{Limitation}

While our proposed framework successfully achieves drift-mitigated trajectory tracking and globally consistent human pose estimation using only a monocular camera, challenges remain regarding physical plausibility during close interactions with the environment.
Specifically, our current method does not explicitly enforce physical constraints between the estimated human mesh and the scene geometry.
Consequently, it sometimes estimate physically implausible motion, which includes unnatural surface penetration or lack of contact.
For instance, in the sitting sequence shown in Fig.~\ref{fig:limit}, although our method reconstructs a posture more consistent with the sitting action and achieves more accurate global placement than the baseline, it still suffers from unnatural surface penetration.
We believe this limitation can be addressed in future work by incorporating scene-aware optimization during the motion inference process. Potential solutions include utilizing Signed Distance Fields (SDF) of the pre-scanned environment to formulate collision-avoidance guidance losses, thereby ensuring physically plausible contact and preventing interpenetration.

\section{Visual Analysis of Trajectory Error in Ablation Study}

\begin{figure}[t]
  \centering
  \includegraphics[width=1.0\linewidth]{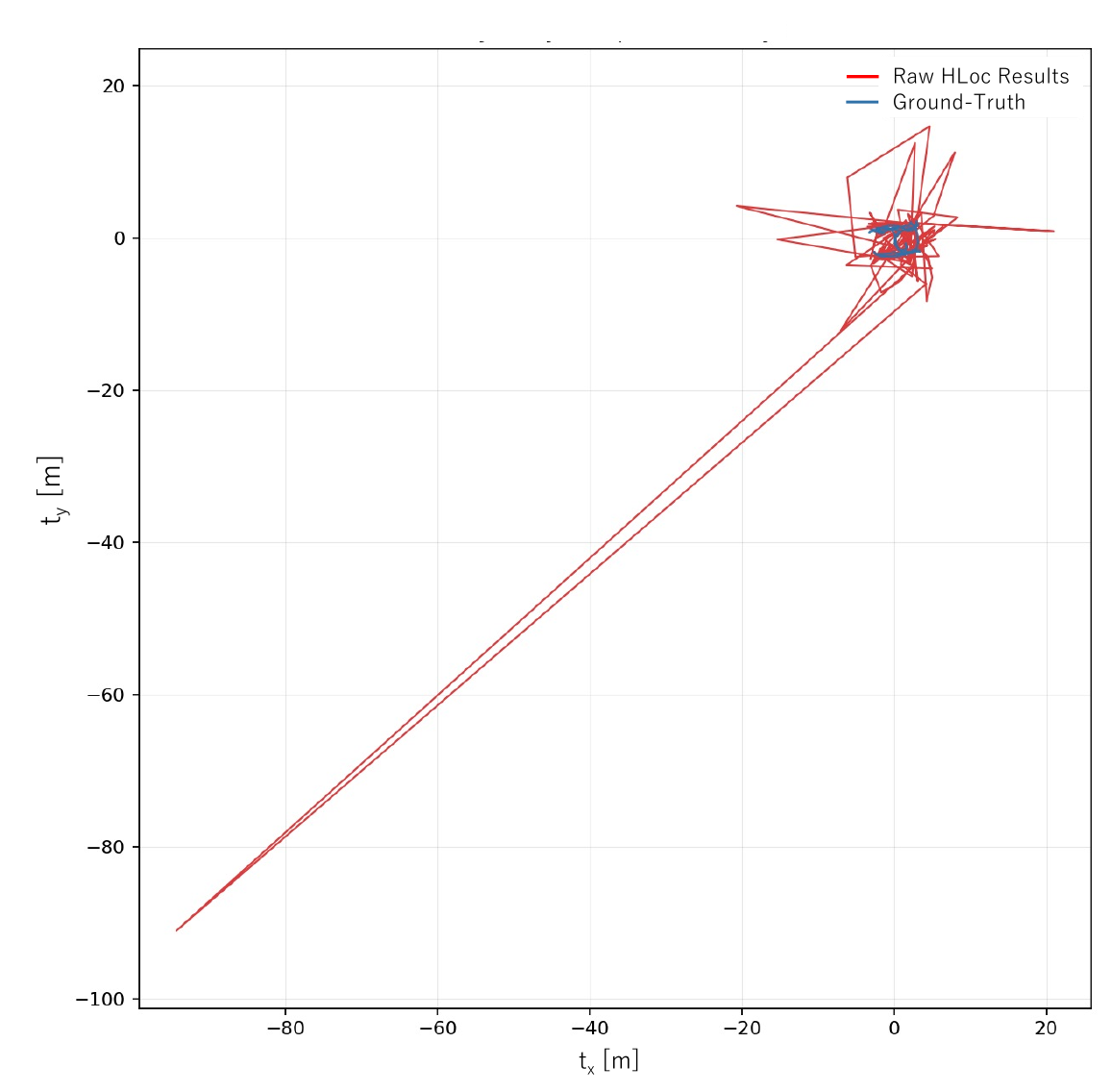}
  \caption{Comparison of global camera trajectories on the horizontal ($t_x$-$t_y$) plane. The blue curve represents the Ground-Truth, while the red segments illustrate the raw HLoc results without our trajectory refinement. }
  \label{fig:ablation}
\end{figure}

To further investigate the necessity of the trajectory refinement \dingc{173}, we visualize the comparison between the ground-truth camera trajectory and the raw trajectory estimated by HLoc on the horizontal ($t_x$-$t_y$) plane in some sequences.
As shown in Fig. ~\ref{fig:ablation}, raw HLoc results frequently deviate by over $10\mathrm{m}$ due to motion blur, creating physically implausible trajectories. These outliers lead to the tracking failure in $\bm{\mathrm{T}_{\text{\textbf{neck}}}}$ and MPJPE observed for Ours w/o \dingc{173} case in Table 2 of the main paper.
Consequently, our trajectory refinement is essential for stable and spatially consistent human pose estimation.
% \section{Video Qualitative Evaluation}

% To conduct more qualitative evaluation of the proposed
% method, we provide a video demo.
% The video demo compares the results of the baseline and our proposed method, Map-Mono-Ego.
% Our method demonstrates results that are closer to the ground truth than the baseline.
% -------------synthetic rendering
% ------------------synthetic rendering

\vfill
\pagebreak

\bibliographystyle{IEEEbib}
\bibliography{refs}

%% file: sections_template/0_abst.tex
Monocular egocentric human pose estimation is essential for ubiquitous activity monitoring.
However, understanding the user's absolute location within the environment remains a challenge.
Existing methods primarily focus on relative motion from an initial position, and tend not to account for the wearer's absolute location within an environment.
Furthermore, inherent scale ambiguity in monocular vision leads to severe translational drift, limiting long-term tracking without specialized multi-sensor hardware.
To address this, we propose \Ours, a novel framework achieving globally consistent human pose estimation solely from a monocular camera by leveraging a pre-scanned 3D point cloud.
We also introduce \Ourdataset, a new dataset pairing egocentric video with ground-truth motion in a scanned environment.
Experiments demonstrate that our approach significantly outperforms the state-of-the-art baseline, proving its utility for practical monitoring tasks without specialized hardware.

%% file: sections_template/1_intro_v2.tex
Estimating human pose using only a lightweight monocular wearable camera, which is common and minimal sensing setting, opens up scalable possibilities for AR/VR and ubiquitous activity monitoring.
% -------------Fig1-------------------------
\begin{figure}[t]
  \centering
  \setlength{\belowcaptionskip}{-15pt}
  \includegraphics[width=0.85\linewidth]{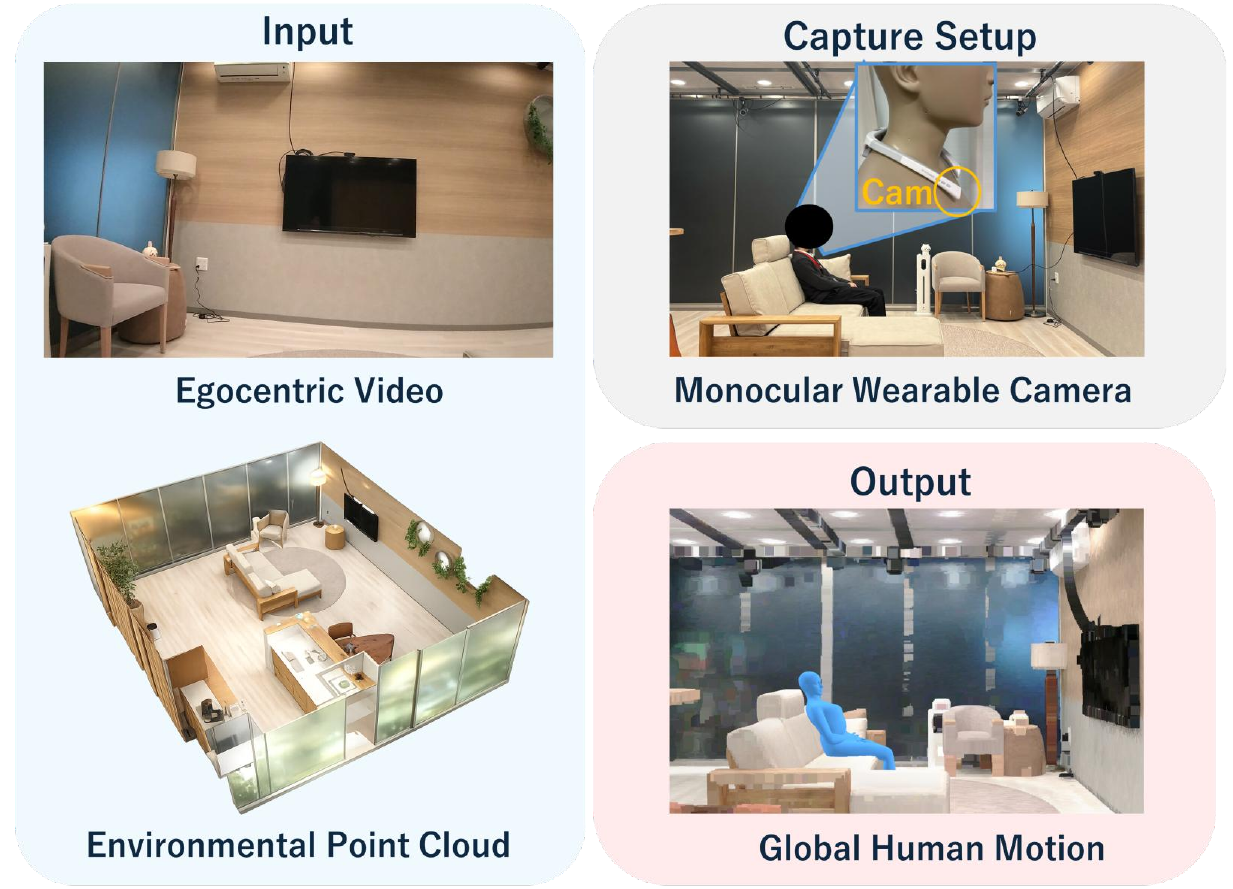}
  \caption{Overview of our proposed \Ours.
  % 3D sceneとかのほうがつたわるかも
  % We take egocentric video from wearable camera and pre-scanned environmental point cloud as input, and estimate human motion in a global scene.
  }
  \label{fig:teaser}
\end{figure}
% -------------Fig1---------------------------
To realize context aware applications, it is essential to understand not only the user’s body posture, but also their spatial relationship with the surrounding environment.
In this paper, we propose \Ours, the framework that achieves spatially consistent human pose estimation from a monocular egocentric camera by leveraging a 3D map pre-scanned by a terrestrial laser scanner as a geometric prior.
Fig.~\ref{fig:teaser} shows the overview of \Ours.

Despite its potential, current egocentric pose estimation methods focus on recovering relative body motion within a local coordinate system, typically initialized at the user's starting position~\cite{egoego,uniegomotion}.
Consequently, these approaches often fail to account for the wearer’s absolute localization within a global map and lack geometric consistency with the environmental structure.
Furthermore, accurate trajectory estimation of commmodity monocular wearable camera is inherently difficult due to scale ambiguity and motion blur, which leads to severe accumulation of translational errors over time~\cite{egoego,slamsurvey}.
While specialized multi-sensor hardware can mitigate these issues, reliance on them limits the applicability.

% While high-end research prototypes, such as Aria Glass~\cite{Ariaglass}, mitigate these issues using sophisticated multi-camera systems and high grade Inertial Measurement Units, such hardware remains rare in consumer grade devices.
% Since most commercial wearable cameras are monocular, reliance on specialized hardware limits the scalability and accessibility of these solutions.

To this end, we focus on monitoring scenarios in controlled environments, such as factories, offices, and residential spaces, where pre-scanning is feasible.
In this paper, we propose a framework that leverages high-density 3D scanned point clouds as a geometric prior for estimating human pose.
By referencing 3D geometry, our framework recovers a drift-mitigated, metric scale camera trajectory from monocular video, enabling precise global pose estimation.
Our contributions are summarized as follows:
1) We propose a framework that estimates globally consistent human pose solely from a monocular egocentric camera by leveraging environmental geometry as a prior.
2) We introduce a robust trajectory tracking algorithm that fuses Hierarchical Localization~\cite{hloc} (HLoc) and Simultaneous Localization and Mapping~\cite{slamsurvey} (SLAM), designed to incorporate environmental priors.
3) We constructed a new benchmark dataset comprising egocentric video, an environmental point cloud, and ground-truth motion data.
Experiments on this dataset demonstrate the effectiveness of our method over the state-of-the-art baseline.
We release the dataset and annotations at \url{https://deguchihiroyuki.github.io/Map-Mono-Ego-Project/}.

% ~\href{https://deguchihiroyuki.github.io/Map-Mono-Ego-Project/}{project page}.
% We will release the dataset and annotations upon acceptance.

%% file: sections_template/2_Related.tex
% \subsection{Egocentric Pose Estimation}
% 各章結びに自身のアプローチの相対化を入れる

% \noindent
% \textbf{Human Motion Estimation from Wearable Devices.}

% \noindent
% \textbf{Human Motion Estimation from Egocentric Video~ } 
\subsection{Human Motion Estimation from Egocentric Video}
Capturing human motion with wearable sensors has gained interest in various fields of application. 
Unlike traditional motion capture systems that consist of multiple external cameras, wearable sensor-based approaches don't require costly equipment and are free from spatial restrictions.
% Capturing human motion with wearable sensors offers a flexible, cost-effective alternative to traditional optical motion capture systems.

%近年では，侵襲性(user fliction)を抑えた実用的な手法を模索する観点からスパースなセンサ構成が注目されており，several IMU configurations, VR trackers, wearable camera devices(downward-facing, outward-facing), Hybrid approaches of IMUs and wearable camerasなどがある．
% Existing methods explore various setups, including sparse IMUs configurations ~\cite{mobileposer, imuposer}, wearable cameras~\cite{egoego, egoallo, uniegomotion}, and hybrid approaches combining IMUs and visual sensors ~\cite{EgoLocate, HPS, avatarposer, egoposer}.
% 近年，正面向きウェアラブルカメラの一人称視点映像のみからのmotion estimationは〜〜という理由からresearch interestを獲得してきている．
Among these, motion estimation solely from egocentric video with a front-facing camera device has gained attention due to its unique capability to capture environmental interactions and the ubiquity of consumer devices.
Since the user's body is often invisible in front-facing views, early approaches relied on interaction cues~\cite{you2me} or control policies~\cite{egopose, kinpoly} to infer pose.
However, a significant paradigm shift occurred with the introduction of EgoEgo~\cite{egoego}, which demonstrated that head movement could be a condition for full-body pose.
Following this, the field has largely shifted toward diffusion-based architectures conditioned on device trajectory~\cite{egoego, egoallo, hmd2, uniegomotion}.
This trend was accelerated by the availability of the standardized devices like Aria Glass~\cite{Ariaglass}, and large-scale datasets containing paired egocentric video and motion ground-truth ~\cite{egosurvey}.
% , such as Nymeria~\cite{nymeria} and Ego-Exo4D~\cite{egoexo4d}.
% NymeriaとかEgoExo4Dは長すぎて引用しづらい．極力避ける
% Consequently, current mainstream approaches predominantly follow this trajectory-conditioned generation pipeline.

Despite these advancements, two critical limitations remain.
First, the performance of these trajectory-conditioned approaches is heavily dependent on the accuracy of the input camera pose~\cite{egoego}.
While high-end devices like Aria Glass provide robust localization, deriving accurate trajectories from standard monocular cameras remains challenging due to scale ambiguity and drift.
% As suggested in EgoEgo~\cite{egoego}, the accuracy of the camera trajectory significantly affects the quality of motion.
Second, prior research has focused on evaluating relative motion starting from an initial position.
Consequently, evaluation and discussion regarding global human placement within the world coordinate system have not been conducted in previous studies.

In this work, we leverage 3D environmental point clouds to mitigate monocular drift and achieve spatially consistent human pose estimation.

% 目的:SLAMとHLocの各特徴を網羅することによる融合の正当化．この章書くのむずそう

\subsection{Camera Pose Estimation}
% \noindent
% \textbf{Wearable Camera Pose Estimation~ }
% Egocentric Human pose estimationを行う上で，Werable Cameraのカメラポーズを時系列的に推定することは基礎的な技術である．
% SLAMやSfMといった技術が基本的にカメラポーズエスティメーションに使われがち．
% しかし，monocularカメラ単体だと，scale ambiguityが解決されない．また，egocentric imageを使う場合だとmotion blurなどによって質の悪い画像が撮影されがちであり，局所特徴量の抽出に失敗してその後のdrift errorなどに繋がったりもする．
% scale ambiguity問題については現在，Arua glassのようなマルチモーダルなセンサを活用することで解決するのが主流となっている．しかし，他の商用デバイスへの拡張性が低い．また，EgoEgoでは学習ベースでオプティカルフローから移動距離を推定するという形でmonocularのscale ambiguityに対して対処したが，依然精度に課題が残る．
% また，SLAMやSfMにおいて，driftを防ぐためLoop Closureが実装されている．しかし，大域的なドリフトを防ぐには十分でなさそう．
% そこで，我々は〜〜という特徴のあるHLocを導入し，大域的なカメラポーズのアンカーとして使用した，
Estimating the 6-DoF camera trajectory is fundamental for egocentric human pose estimation.
While SLAM is traditionally adopted for this purpose, relying solely on a monocular camera still primarily suffers from scale ambiguity and accumulated drift caused by rapid motion and blur~\cite{slamsurvey}.
% Traditionally, geometric approaches such as Simultaneous Localization and Mapping (SLAM) and Structure from Motion (SfM) have been widely adopted for this purpose.
% However, relying solely on a monocular camera introduces inherent challenges, primarily the scale ambiguity problem.
% Furthermore, egocentric videos often suffer from severe motion blur, leading to tracking failures and significant accumulated drift errors.

To address scale ambiguity, recent approaches often leverage multi-modal sensors like Aria Glass~\cite{Ariaglass}.
While effective, relying on specialized hardware limits the scalability to consumer-grade devices.
Alternatively, EgoEgo ~\cite{egoego} attempts to resolve scale from monocular video by learning the relationship between distance scale and optical flow.
However, these methods often struggle to maintain geometric consistency over long sequences.
Although standard SLAM systems incorporate loop closure to mitigate drift, they are often insufficient in scenarios where the user does not revisit previous locations frequently or in texture-less environments~\cite{slamsurvey}.

% To address this, we integrate Hierarchical Localization (HLoc)~\cite{hloc}, which employs a coarse-to-fine pipeline to establish robust 2D-3D correspondences between the query image and the pre-built 3D map. By leveraging the resulting absolute poses as global anchors to constrain the SLAM trajectory, our method ensures drift-free long-term tracking using only a monocular camera.

To address these limitations, we integrate HLoc~\cite{hloc} into the pipeline.
Unlike pure SLAM or learning-based odometry, HLoc, which is a structure-based localization method~\cite{structure-base}, leverages a pre-built 3D map to establish robust 2D-3D correspondences. 
By using these absolute poses as global anchors, our method ensures drift-mitigated, metric scale tracking using only a monocular camera, providing a reliable foundation for human motion estimation.

%% file: sections_template/3_method_v2.tex
\begin{figure*}[t]
  \centering
  \setlength{\belowcaptionskip}{-8pt}
  \includegraphics[width=0.97\linewidth]{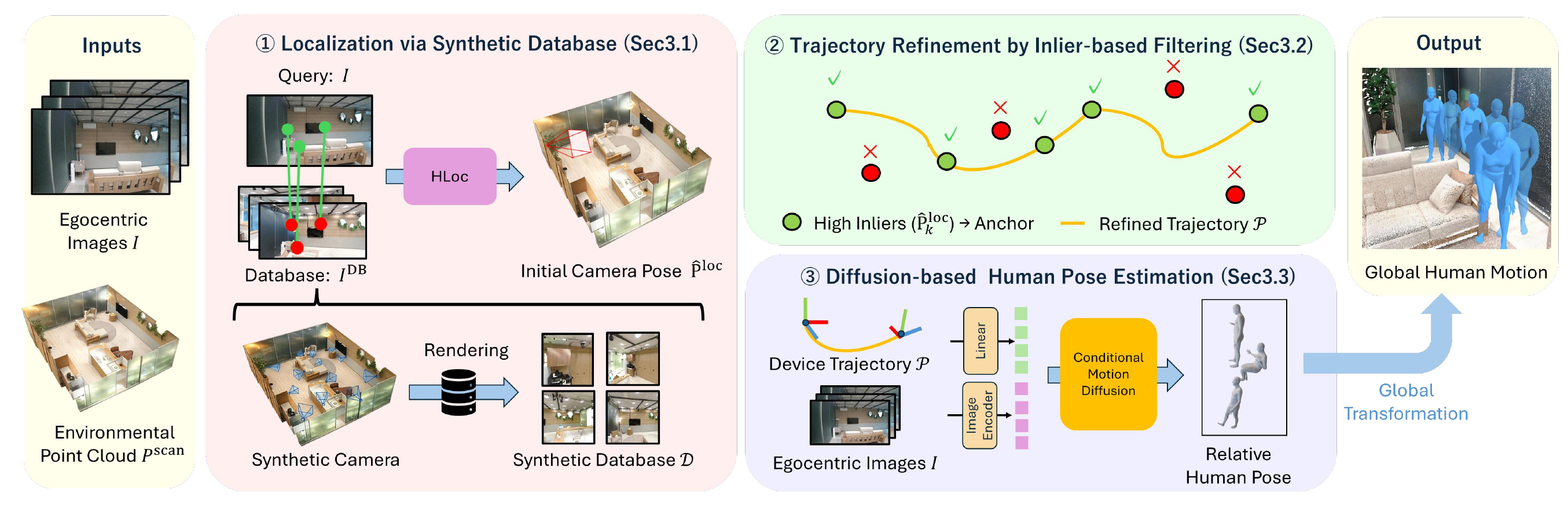}
  \caption{Overview of the ~\Ours{} framework. \dingc{172} We estimate initial camera pose $\hat{\mathbf{P}}^{\text{loc}}$ (Sec.~\ref{subsec:syntheticLoc}). \dingc{173} We filter $\hat{\mathbf{P}}^{\text{loc}}$ for reliable camera poses and recover a smooth, drift-free trajectory $\mathcal{P}$ via SLAM-based interpolation (Sec.~\ref{subsec:camera trajectory estimation}).\\
  \dingc{174} We predict global human motion using the refined trajectory. $\mathcal{P}$ (Sec.~\ref{subsec:humanpose}).}
  \label{fig:overview}
\end{figure*}
Our goal is to recover the global human motion sequence $X$ from $T$ frames of an egocentric video $I = \{I_t\}_{t=1}^T$, and a pre-scanned 3D point cloud $P^{scan}$.
As illustrated in Fig.~\ref{fig:overview}, \Ours{} operates in three stages:
\textbf{\dingc{172} Localization via Synthetic Database: } Estimating camera poses initially by matching the video frames against a synthetically rendered database from the point cloud.
\textbf{\dingc{173} Trajectory Refinement by Inlier-based Filtering: } Filtering for reliable camera poses based on geometric consistency and interpolating the trajectory using SLAM to ensure smoothness and global consistency.
\textbf{\dingc{174} Diffusion-based Human Pose Estimation: } Predicting the human motion sequence using a diffusion model conditioned on the refined trajectory.
By feeding the robust trajectory from \dingc{172} and \dingc{173} into the motion diffusion pipeline of \dingc{174}, our framework achieves spatially consistent monocular egocentric human pose estimation.

\subsection{Localization via Synthetic Database}
\label{subsec:syntheticLoc}

To estimate the camera poses within the scanned environment, we initially employ HLoc~\cite{hloc}.
Since HLoc requires an image-to-geometry reference, we construct $N$ frames of synthetic database $\mathcal{D} = \{ I_n^{\text{db}}, \mathbf{P}_{n}^{\text{db}}, \mathcal{C}_n \}_{n=1}^{N}$ from $P^{scan}$.
Specifically, we generate synthetic views and ground-truth poses from $P^{scan}$ following the protocol in HPS~\cite{HPS}.
Each entry contains the synthetic image $I_n^{\text{db}}$, its pose $\mathbf{P}_{n}^{\text{db}} \in \mathrm{SE(3)}$, and 3D correspondences $\mathcal{C}_n$.
% Specifically, we sample virtual cameras within $P^{scan}$ at regular spatial intervals with random orientations and render synthetic views following the protocol in HPS~\cite{HPS}.
% Each entry contains the synthetic image $I_n^{\text{db}}$, its ground-truth camera pose $\mathbf{P}_{n}^{\text{db}} \in \mathrm{SE(3)}$, and exact 2D-3D correspondences $\mathcal{C}_n$.
We then treat the input video frames $\mathcal{I}$ as queries.
By performing visual localization against $\mathcal{D}$, we obtain the initial camera pose $\hat{\mathbf{P}}_{t}^{\text{loc}} \in \mathrm{SE(3)}$ for each frame $t$.

% ------------fig3-------------
\begin{figure}[t]
  \centering
  \setlength{\belowcaptionskip}{-10pt}
  \includegraphics[width=1.0\linewidth]{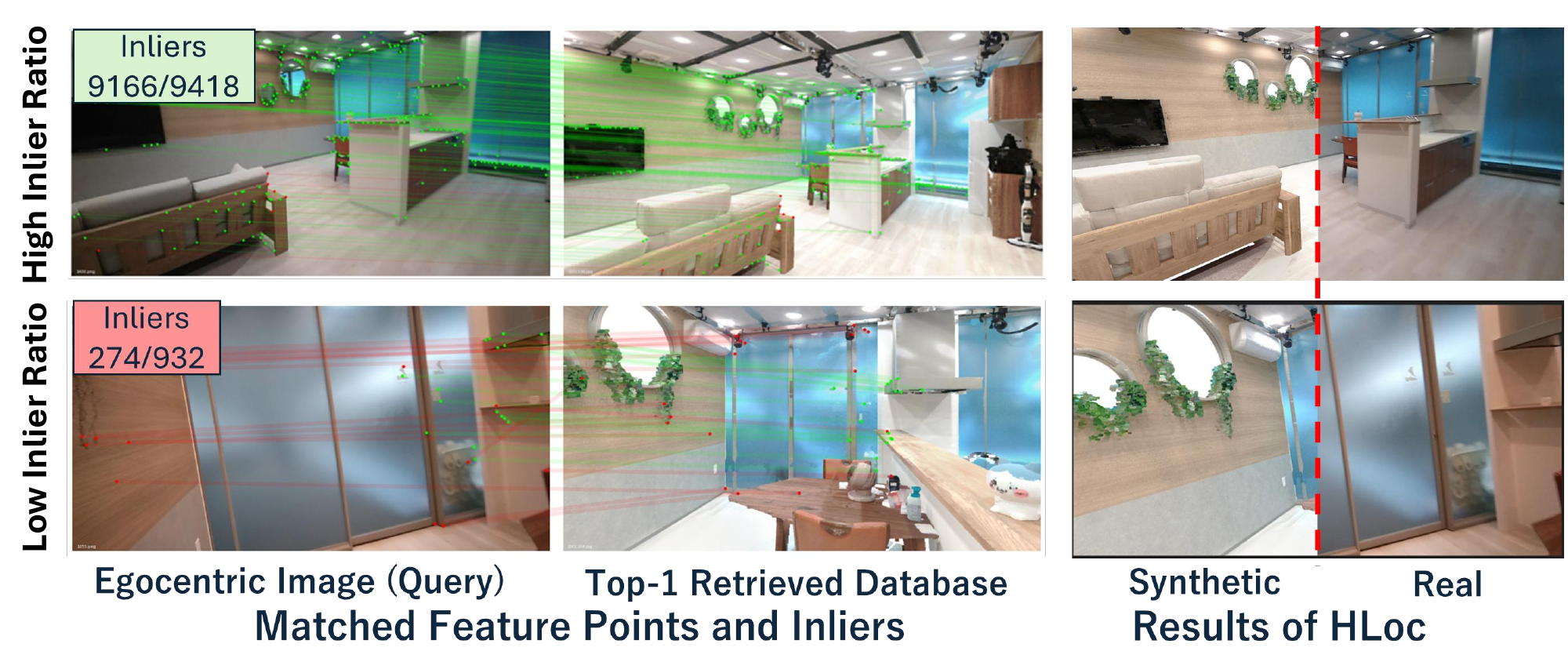}
  \caption{The left columns visualize matches between egocentric images and top-1 retrieved images. The right columns compare real egocentric views with synthetic views, which render point cloud $P^{\text{scan}}$ from estimated poses. High inlier metrics (top) demonstrate precise alignment, while low metrics (bottom) reveal noticeable misalignment.}
  \label{fig:inlier}
\end{figure}
% -------------fig3-------------

\subsection{Trajectory Refinement by Inlier-based Filtering}
\label{subsec:camera trajectory estimation}

The raw poses $\hat{\mathbf{P}}_{t}^{\text{loc}}$ may contain outliers due to motion blur or textureless regions.
% To ensure reliability, we filter these poses based on the PnP inlier ratio, which correlates strongly with localization accuracy (see Fig.~\ref{fig:inlier}).
% We retain only the poses where the inlier ratio exceeds a specific threshold, defining a subset of reliable indices $\mathcal{K} \subset \{1, \dots, T\}$.
To ensure reliability, we filter these poses based on the PnP inlier count and ratio derived from aggregated matches against the top-40 database candidates retrieved by a global descriptor in HLoc.
As visualized in Fig.~\ref{fig:inlier}, these metrics correlate strongly with localization accuracy.
We retain only the poses where both the inlier count and ratio exceed specific thresholds, defining a subset of reliable indices $\mathcal{K} \subset \{1, \dots, T\}$.

To reconstruct the full trajectory $\mathcal{P} = \{\mathbf{P}_t\}_{t=1}^T$, we interpolate between the reliable anchor poses $\{ \hat{\mathbf{P}}_{k}^{\text{loc}} \}_{k \in \mathcal{K}}$ by aligning the continuous trajectory estimated in a local coordinate frame by DROID-SLAM~\cite{teed2021droid}.
Consider an interval between two consecutive reliable frames $n, m \in \mathcal{K}$ ($n < m$) the corresponding SLAM pose sequence $\{ \mathbf{P}^{\text{slam}}_t \}_{t=n}^{m}$.
First, we compute the similarity transformation $S \in \mathrm{Sim}(3)$ that aligns the SLAM defined frame to the world frame at the start frame $n$:
\begin{equation}
    S = \hat{\mathbf{P}}_{n}^{\text{loc}} (\mathbf{P}^{\text{slam}}_n)^{-1}.
\end{equation}
To correct the accumulated scale drift inherent in monocular SLAM, we define a residual transformation $E \in \mathrm{Sim}(3)$ that aligns the propagated pose with the reliable anchor at the end frame $m$:
\begin{equation}
    E = \hat{\mathbf{P}}_{m}^{\text{loc}} (S \mathbf{P}^{\text{slam}}_m)^{-1}.
\end{equation}
We distribute this residual across the interval using a time-dependent factor $\alpha_t = \frac{t-n}{m-n}$.
Finally, the refined global camera pose $\mathbf{P}_t$ is computed via Lie algebra interpolation~\cite{sim(3)}:
\begin{equation}
    % \mathbf{P}_t = \text{Exp}(\alpha_t \text{Log}(E)) S \mathbf{P}^{\text{slam}}_t.
    \mathbf{P}_t = \exp(\alpha_t \log(E)) S \mathbf{P}^{\text{slam}}_t.
\end{equation}
This formulation ensures the trajectory strictly satisfies boundary constraints at $n$ and $m$ while preserving the local geometric structure captured by SLAM.

\subsection{Diffusion-based Human Pose Estimation}
\label{subsec:humanpose}

Finally, we estimate the SMPL-X~\cite{smplx} motion sequence $X = \{\{\mathrm{R}^{\text{root}}_t, \mathrm{t}^{\text{root}}_t, \theta_t \}_{t=1}^T, \beta\}$ using a Transformer-based diffusion model following UniEgoMotion~\cite{uniegomotion}.
Here, $\mathrm{R}^{\text{root}}_t \in \mathbb{R}^{3}$ and $\mathrm{t}^{\text{root}}_t \in \mathbb{R}^{3}$ denote the root joint's global rotation and translation, $\theta_t \in \mathbb{R}^ {21 \times 3}$ denotes the local joint angles excluding hands and face, and $\beta \in \mathbb{R}^{10}$ represents the time invariant body shape.
Unlike UniEgoMotion~\cite{uniegomotion}, which targets head-mounted devices, we use a neck-mounted camera (as shown in Fig.~\ref{fig:teaser})~\cite{thinklet} in our experiment.
Therefore, we retrained the model to accept the neck joint trajectory as the condition.

% The model is conditioned on the refined camera trajectory $\mathcal{P}$ and image features extracted by DINOv2~\cite{dinov2}.
Before the inference, we transform the refined camera trajectory $\mathcal{P}$ into a canonicalized trajectory $\mathcal{P}^{cano}$ where the first frame is centered at the origin and aligned with the forward axis.
The architecture iteratively recovers $X$ from Gaussian noise, conditioned on the canonicalized camera trajectory $\mathcal{P}^{cano}$ and DINOv2~\cite{dinov2} image features.
Finally, we transform the predicted relative motion back to the world frame by applying the inverse canonicalization to $\mathrm{R}^{\text{root}}$ and $\mathrm{t}^{\text{root}}$.

%% file: sections_template/4_experiments.tex
\subsection{Dataset}
% \noindent
% \textbf{Dataset. } 
% Diffusion部分Trainingは既存のEE4D motion dataset使った．
% 10FPS80frameのシーケンスで実験を行った
% ベンチマーク用に評価に関しては自分たちでデータを撮った．
% 一人称視点画像はthinkletで撮影
% FARO使って点群とった
% GT poseはmarlerless mocap（Acuity?）を使ってまず姿勢を取得したのち,まずDB形式のマーカーmodelに変換，その後最後にsomaを通してSMPL modelを出力．
To train the motion diffusion model, we use EE4D-motion dataset ~\cite{uniegomotion}.
Following UniEgoMotion~\cite{uniegomotion}, we trained on 8-second videos at 10fps. 
On the other hand, for benchmarking, a dataset pairing environmental point clouds, egocentric video, and ground-truth motion data was required.
Therefore, we constructed \Ourdataset.
We collected the data in a laboratory environment designed to simulate a typical living room, assuming a real-world scenario of daily activity monitoring.
\Ourdataset{} comprises 152 sequences of 8 seconds each at 10 FPS, and we used it for evaluation.
For more details, please refer to the supplementary material.
% この辺りの詳細 supplemental行きがまるそう

% リビングを模した環境で日常的な動作を撮影した
% \noindent
% \textbf{Implementation Details. }

% \noindent
% \textbf{Baseline. }
\subsection{Baseline}
% monocularっていうのがあんまやられてないていうのをここでも強調しておくといいかも
% introとかでやられているAriaと比較するような導入をここでもいれておくとよさそう
% To our knowledge
% Recent methods~\cite{egoallo, hmd2, uniegomotion} often rely on high-end sensors like Aria glasses~\cite{Ariaglass} for accurate carera trajectory tracking, limiting scalability.
% In contrast, the monocular setting is crucial for deployment but remains under-explored due to the difficulty of single-camera tracking.
% To our knowledge, no existing approach addresses global human pose estimation utilizing a monocular egocentric video and a pre-scanned environmental point cloud.
% Consequently, to validate the effectiveness of incorporating environmental priors, we compare our approach against a method that relies solely on monocular visual data.
% We adopt the trajectory estimation algorithm from EgoEgo~\cite{egoego}, the current state-of-the-art method specifically designed for monocular egocentric human pose estimation, and refer to it as our baseline.
To validate the benefit of environmental priors within the scalable monocular setting (avoiding reliance on high-end sensors like Aria glass~\cite{Ariaglass, egoallo, hmd2}), we compare against a method relying solely on monocular vision.
We adopt the trajectory estimation algorithm from EgoEgo~\cite{egoego}, the state-of-the-art method specifically designed for the monocular egocentric human pose estimation, and refer to it as our baseline.

EgoEgo employs a hybrid approach that integrates learning-based algorithms with monocular SLAM to achieve metric scale camera trajectory estimation.
Specifically, it combines DROID-SLAM~\cite{teed2021droid} for trajectory tracking with GravityNet for gravity alignment and HeadNet, which processes optical flow features extracted by RAFT~\cite{raft} and ResNet-18~\cite{resnet}, to estimate the metric scale translation and rotation of a monocular camera.
In our implementation, we utilize the official pretrained models for both GravityNet and HeadNet.
To isolate the impact of the trajectory estimation method, we feed the metric-scale trajectory obtained by this EgoEgo-based framework into the same motion diffusion model used in \Ours.
Furthermore, since the baseline operates in a local coordinate system relative to the start frame, we align its initial global position and orientation using the results of our method to ensure a fair comparison.

%---------table
\input{tabs/quantiative}
% ------table

\subsection{Evaluation Metrics}
To assess neck-mounted camera tracking accuracy, we report the Neck Orientation Error ($\bm{\mathrm{O}_{\text{\textbf{neck}}}}$) and the Translation Error ($\bm{\mathrm{T}_{\text{\textbf{neck}}}}$), measured in $\mathrm{mm}$. The orientation error is computed as the Frobenius norm of the rotation matrix.
For pose accuracy, we report standard protocols, all measured in $\mathrm{mm}$.
\textbf{MPJPE} computes the mean per-joint positional error over 22 body joints.
\textbf{MPJPE-Rigid} further aligns the predicted motion to the ground truth using a single rigid transformation \textit{per sequence} to remove a global offset, and measures \textit{sequence-level} motion consistency.
\textbf{MPJPE-PA} applies Procrustes analysis \textit{per frame} to align the predicted and ground-truth motions, measuring the accuracy of \textit{frame-level} local pose predictions.
To evaluate physical realism, we report Foot Sliding (\textbf{FS})~\cite{nemf} and Foot Contact (\textbf{FC}), both measured in $\mathrm{mm}$. Specifically, FS quantifies the sliding distance when the foot is close to the ground, while FC computes the average foot–ground separation to capture floating and penetration artifacts.
% Finally, we introduce Semantic Similarity (\textbf{SS}) to assess motion quality in a semantic latent space. Specifically, leveraging the motion encoder from TMR~\cite{tmr}, we compute the cosine distance between the embeddings of the predicted and ground-truth motions, providing a perceptual evaluation of motion similarity.
Finally, we report Semantic Similarity (\textbf{SS}), which evaluates perceptual motion quality in a semantic latent space via the TMR~\cite{tmr} motion encoder.

\begin{figure}[t]
  \centering
  \setlength{\belowcaptionskip}{-10pt}
  \includegraphics[width=1.0\linewidth]{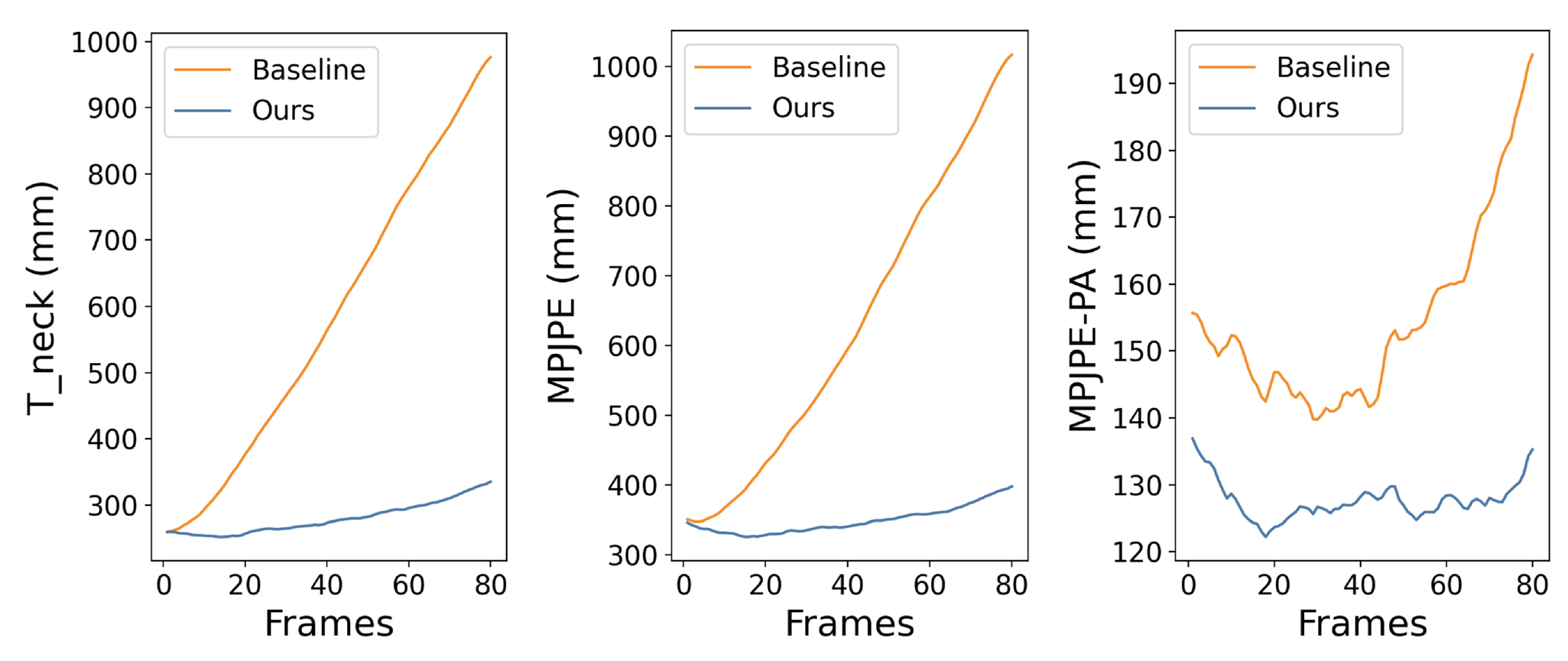}
  \vspace{-15pt}
  \caption{We plot the average $\bm{\mathrm{T}_{\text{neck}}}$, MPJPE, and MPJPE-PA errors at each frame index across all sequences. Our method achieves drift-mitigated human motion estimation.}
  \label{fig:temporal}
\end{figure}

\subsection{Comparison with the Baseline}

As summarized in Table~\ref{tab:quantitative}, Fig.~\ref{fig:qualitative}, and Fig.~\ref{fig:temporal}, we evaluate our pipeline against the baseline on \Ourdataset.

% --- Table 1 Analysis (大幅加筆部分) ---
As shown in Table~\ref{tab:quantitative}, our method achieves significantly more accurate results in $\mathrm{T}_{\text{neck}}$ and competitive performance in $\mathrm{O}_{\text{neck}}$. 
This confirms that our map-grounded approach estimates a more precise camera trajectory than the baseline.
Crucially, our method consistently outperforms the baseline across all remaining pose metrics, including MPJPE-Rigid, MPJPE-PA, and Semantic Similarity (SS).
The poor performance of the baseline on these metrics suggests a fundamental issue in the motion estimation process.
We attribute this to the failure of the baseline in estimating the correct metric scale and gravity alignment.
Since the diffusion model is conditioned on the neck trajectory, an input trajectory with incorrect translational scale or gravity alignment acts as a physically inconsistent condition.
It makes the motion diffusion model output an unnatural body posture, thereby degrading not only global positioning but also the local fidelity and semantic quality of the estimated motion.

% --- Qualitative Results (Fig 3) ---
Qualitative comparisons in Fig.~\ref{fig:qualitative} further validate these findings.
Our method successfully achieves precise global human motion, accurately reconstructing interactions with the environment, such as sitting near a robot vacuum or reaching for a microwave.
In contrast, the baseline predicts unnatural motions characterized by significant translational drift and erroneous neck heights.
It shows that monocular methods struggle to estimate plausible human-scene interaction without environmental priors.
% 重複を避けるため、ここは「現象（浮いている、ズレている）」の記述に留め、原因（Scale/Gravity）は上のパラグラフで述べたので繰り返さない
% These artifacts visually confirm that without environmental priors, monocular methods struggle to maintain the geometric consistency required for plausible human-scene interaction.

% --- Temporal Stability (Fig 5) ---
% Furthermore, to analyze temporal stability, Fig.~\ref{fig:temporal} illustrates the frame-wise evaluation of average errors across all test sequences.
Furthermore, frame-wise analysis in Fig.~\ref{fig:temporal} clearly demonstrates that the baseline suffers from a severe accumulation of errors in $\mathrm{T}_{\text{neck}}$ and MPJPE over time, exhibiting the characteristic effect of monocular drift.
Notably, the baseline's local pose fidelity (MPJPE-PA) also degrades in later frames, whereas ours remains stable.
% This reinforces our observation that accumulated drift in the camera trajectory leads to increasingly unnatural conditioning, causing the diffusion model to predict implausible poses over long durations.
This stability highlights how robust trajectory tracking prevents the diffusion model from generating implausible poses over long sequences.

% 図の高さはんぶんくらいにできたらよさそう

\input{tabs/ablation}

\subsection{Ablation Study}
We studied the effects of our trajectory refinement using several metrics in Table~\ref{tab:ablation}. 
Without this refinement, raw HLoc localization suffers from numerous outliers, leading to inaccurate and discontinuous camera trajectories.
It leads to a substantial increase in MPJPE, demonstrating the necessity of our refinement process for stable global pose estimation.
% Please refer to the supplementary material for more experimental results.

%% file: tabs/quantiative.tex
\begin{table*}[tb]
\centering
\setlength{\belowcaptionskip}{-6pt}
\caption{Quantitative comparison. Bold numbers denote the better performance for each metric.}
\vspace{-6pt}
\scalebox{1.0}[1.0]

% \vspace{0.5em}
{
\begin{tabular}{c||ccccccccc}
\hline & $\mathrm{O}_{\text{neck}}\downarrow$ & $\mathrm{T}_{\text{neck}}\downarrow$ 
 & MPJPE$\downarrow$ & MPJPE-Rigid$\downarrow$ & MPJPE-PA$\downarrow$ &
 FS$\downarrow$ & FC$\downarrow$ & SS$\uparrow$ & \\ \hline
Baseline & $\mathbf{0.66}$ & $577.3$ & $623.0$ & $351.0$& $152.9$    & $16.0$ & $7.8$ & $0.76$ \\
            Ours & $\mathbf{0.66}$ & $\mathbf{279.7}$ & $\mathbf{347.5}$ & $\mathbf{201.3}$ & $\mathbf{128.6}$ & $\mathbf{11.4}$ & $\mathbf{2.2}$  & $\mathbf{0.82}$  
            \\ \hline
\end{tabular}}
\label{tab:quantitative}
\end{table*}

%% file: tabs/ablation.tex
\begin{table}[tb]
\centering
\caption{Ablation study for the effects of our trajectory refinement \dingc{173}. Values denoted by '---' indicate that the error exceeded $10^4\mathrm{mm}$, representing a failure in the estimation.}
\vspace{-6pt}
\scalebox{1.0}[1.0]

% \vspace{0.5em}
{
\begin{tabular}{c||cccc}
\hline & $\mathrm{O}_{\text{neck}}\downarrow$ & $\mathrm{T}_{\text{neck}}\downarrow$ 
 & MPJPE$\downarrow$  \\ \hline
Ours & $\mathbf{0.66}$ & $\mathbf{279.7}$ & $\mathbf{347.5}$ \\
            Ours w/o \dingc{173}& $1.41$ & --- & --- 
            \\ \hline
\end{tabular}}
\label{tab:ablation}
\end{table}

%% file: sections_template/5_conclusion.tex
In this study, we propose \Ours, the framework that effectively utilizes environmental point clouds and monocular egocentric video to estimate the global human pose.
Specifically, we leverage environmental point clouds as geometric priors through HLoc-based localization and inlier-based trajectory refinement.
By integrating this robust tracking into a diffusion framework, we realize spatially consistent motion estimation.
Experiments demonstrate that our method significantly outperforms the state-of-the-art baseline method, and shows its utility for practical monitoring applications in daily environments.
For future work, we aim to explicitly incorporate the 3D scene geometry into the motion inference process to predict more physically plausible motions, including plausible human-object and human-scene interactions.

% We hope that our ~\Ours will contribute to further developments in this research field.

\vspace{4pt}
% TO DO FIX
\noindent
\textbf{Acknowledgement~ } 
This work was supported by JST BOOST, Japan Grant Number JPMJBS2409.
This work was also supported by Council for Science, Technology and Innovation, “Cross-ministerial Strategic Innovation Promotion Program (SIP), Development of foundational technologies and rules for expansion of the virtual economy" (JPJ012495). (funding agency: NEDO).